\documentclass{ieeetran}

\let\labelindent\relax
\usepackage{graphicx}
\usepackage{pifont}
\usepackage{subcaption}
\usepackage{eqlist}
\usepackage{amsfonts}
\usepackage{tabularx}
\usepackage{booktabs}
\usepackage{amssymb}
\usepackage{amsmath}
\usepackage{enumitem}
\usepackage{threeparttable}
\usepackage[lined, ruled, linesnumbered, commentsnumbered]{algorithm2e}
\usepackage{lipsum}
\usepackage{dashrule}
\usepackage[
  colorlinks,
  linkcolor=magenta
]{hyperref}
\newcommand\rurl[1]{%
  \href{http://#1}{\nolinkurl{#1}}%
}

\usepackage[usenames,dvipsnames]{xcolor}
\usepackage{comment}

\definecolor{cpurple}{rgb}{0.93,0.098,0.584}
\usepackage{multirow}

\usepackage{url}
\usepackage{CJKutf8} 
\definecolor{bleudefrance}{rgb}{0.19, 0.55, 0.91}
\definecolor{awesome}{rgb}{1.0, 0.13, 0.32}

\usepackage[normalem]{ulem}

\begin{document}
\title{Automatically Prepare Training Data for YOLO Using Robotic In-Hand Observation and Synthesis}

\author{Hao Chen$^{1}$, Weiwei Wan$^{1*}$, Masaki Matsushita$^{2}$, Takeyuki Kotaka$^{2}$ and Kensuke Harada$^{13}$
\thanks{$^{1}$Department of System Innovation, Graduate School of Engineering Science, Osaka University, Toyonaka, Osaka, Japan.}
\thanks{$^{2}$H.U. Group Research Inst. G. K., Japan.}
\thanks{$^{3}$National Inst. of AIST, Japan.}
\thanks{$^{*}$Contact: Weiwei Wan, {\tt\small wan@sys.es.osaka-u.ac.jp}}}

\markboth{Arxiv Version, 2022}
{Chen \MakeLowercase{et al.}: Precisely Recognizing and Locating Test Tubes in A Rack Using Structured-Light Depth Sensing} 

\maketitle

\begin{abstract}
Deep learning methods have recently exhibited impressive performance in object detection. However, such methods needed much training data to achieve high recognition accuracy, which was time-consuming and required considerable manual work like labeling images. In this paper, we automatically prepare training data using robots. Considering the low efficiency and high energy consumption in robot motion, we proposed combining robotic in-hand observation and data synthesis to enlarge the limited data set collected by the robot. We first used a robot with a depth sensor to collect images of objects held in the robot's hands and segment the object pictures. Then, we used a copy-paste method to synthesize the segmented objects with rack backgrounds. The collected and synthetic images are combined to train a deep detection neural network. We conducted experiments to compare YOLOv5x detectors trained with images collected using the proposed method and several other methods. The results showed that combined observation and synthetic images led to comparable performance to manual data preparation. They provided a good guide on optimizing data configurations and parameter settings for training detectors. The proposed method required only a single process and was a low-cost way to produce the combined data. Interested readers may find the data sets and trained models from the following GitHub repository: \rurl{github.com/wrslab/tubedet}

\def\abstractname{Note to Practitioners}
\begin{abstract}
The background of this study is a requirement in lab automation -- Using robots to arrange randomly placed tubes automatically. Before sending test tubes to an examination machine for gradient tests, humans need to categorize and organize the tubes into specific patterns to fit the machine's internal design. Employing humans is difficult as the tube arrangement requirements are time-varying. A preferred solution is using robots to replace humans. The robots should have a vision system to detect the tubes and a manipulation system to perform physical arranging actions. They will be used in busy seasons while deployed for other tasks in leisure time. Deep neural networks like YOLO are effective for the tube detection task. However, preparing the training data is challenging and unsuitable for lab end users. Pre-trained neural networks are options but have limited tube detection ability and cannot deal with newly included tube types. The method developed in this work helps solve the training data preparation problem. With its support, the robot can automatically prepare training data that has comparable quality to manually labeled ones in a single-process and low-cost way.
\end{abstract}
\end{abstract} 

\begin{IEEEkeywords}
Robotic data preparation, data synthesis, test tube detection
\end{IEEEkeywords}

\section{Introduction}
\label{sec:introduction}

\begin{figure}[!t]
    \centerline{\includegraphics[width=\linewidth]{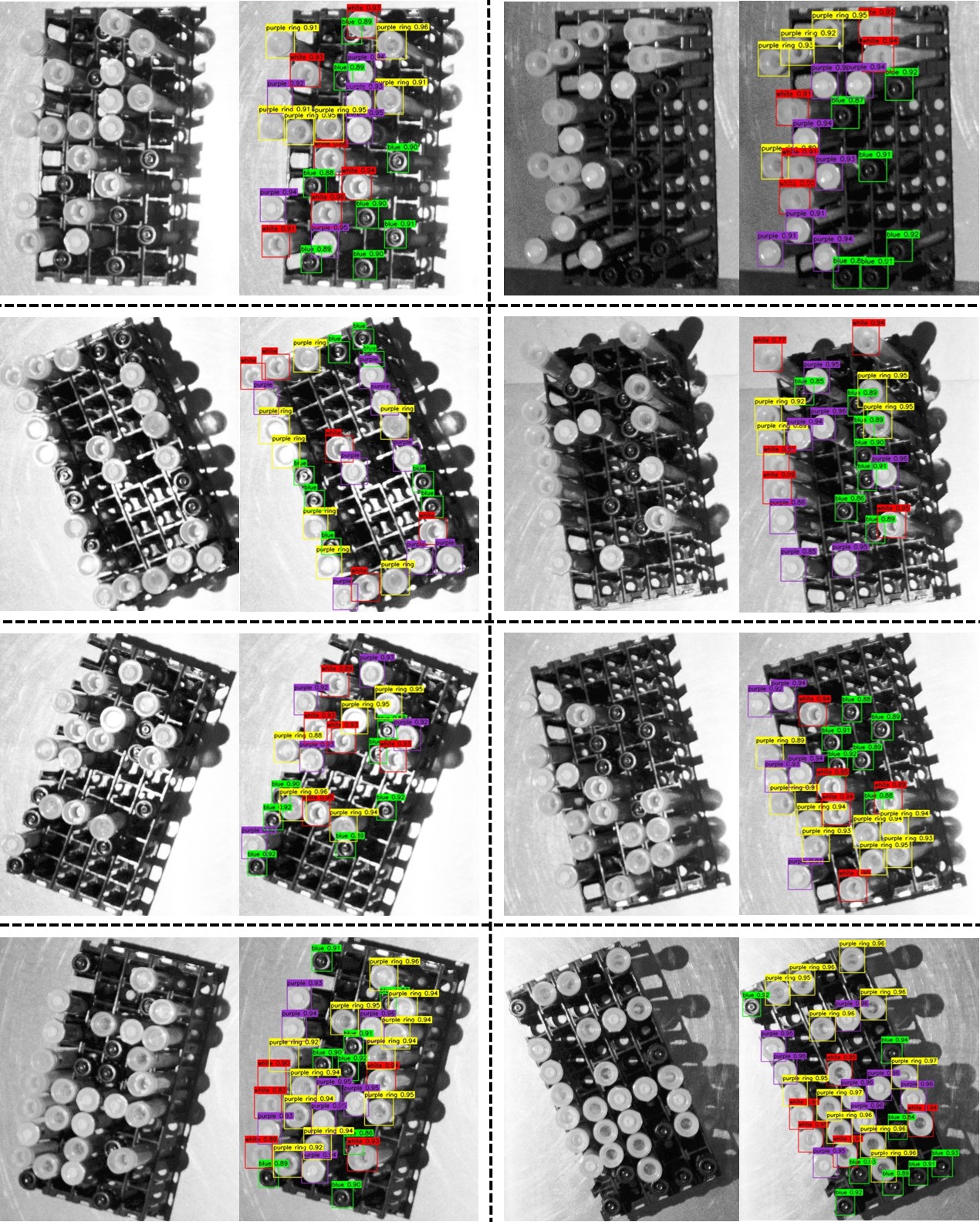}}
    \caption{Several examples of in-rack test tube detection. Each grid includes two images. The left image is captured by a vision sensor. The right image is the recognition result. The data used for training the detection neural network is prepared using the proposed method.}
    \label{fig:in_rack_test_tube_detection}
\end{figure}
Recent advances in deep learning have led to a revolution in object detection. Deep learning-based methods use deep neural networks to learn features from training data. They outperform traditional hand-crafted features with impressive results. Despite these advantages, deep learning-based object detection requires collecting a large amount of labeled data for training, which is time-consuming and labor-intensive, and has significantly hindered the scalability and flexibility of deep learning-based applications.

Previously, researchers have developed several methods to reduce data collection costs. For example, data augmentation \cite{shorten2019survey} enriched existing training data sets by applying random transformations like image rotation or scaling. Data synthesis \cite{gupta2016synthetic}\cite{yi2019generative} generated previously unseen data using simulation or adversarial neural networks. The main challenge of the augmentation or synthesis methods was the ``domain gap'' \cite{hinterstoisser2019an}\cite{karsch2011rendering}: Augmented data had less varied visual contexts. Synthesized data was prone to discrepancies with the real world. Recently, researchers have revisited using the copy-paste method \cite{perez2003poisson} to increase data. The method was effective in compensating for the ``domain gap'' problem, exhibiting impressive performance. There is no clear boundary between augmentation and synthesis when using the copy-paste method to generate data. It was mainly classified as an synthesis method \cite{dwibedi2017cut}\cite{georgakis2017synthesizing}, although some studies considered it to be augmentation \cite{ghiasi2021simple}. This paper calls it a synthesis method to avoid confusion with transformation and scale-based data generation.

The most tiring aspect of the copy-and-paste method is how to neatly cut a large variety of target regions and paste them onto a new background. Previously, researchers working on robotic manipulation have developed robotic methods to segment novel objects from backgrounds. For example, Florence et al. \cite{florence2020robot}, Boerdijk et al. \cite{boerdijk2021s}, and Pathak et al. \cite{pathak2018learning} respectively used robotic in-hand or non-prehensile manipulation to change objects' observation viewpoints and segmented the objects based on the robot motion. Such systems could replace humans to segment goal object regions under various conditions. Very recent studies \cite{boerdijk2021s}\cite{boerdijk2020self} has noticed the advantage, and increased data size and contextual variety by pasting the objects segmented by robotic systems onto random backgrounds. Despite their seminal proposals, the need for copy-and-paste synthesis and the impact of data volume and ratios remain undiscussed.

Based on the current research status, this paper further delves into using robots to collect training data automatically. Considering the low efficiency and high energy consumption in robotic data collection, we propose combining robotic observation and copy-paste synthesis to reduce costs. We assume a test tube detection task shown in Fig. \ref{fig:in_rack_test_tube_detection} and use a robot with a depth sensor to move and observe tubes. The robot collects observation images and, at the same time, segments tubes from the images for copy-paste synthesis. The observation and synthetic images are used as training data for deep detection neural networks. Especially for the synthesis routine, we value the co-occurrence of tubes and racks, and paste tubes inside a rack area to obtain contextual consistency. Also, we take into account factors like tube-to-tube occlusions and foreground changes caused by environment or visual difference to reduce unrealistic synthetic results. The proposed method helps enrich the data set and resolve the ``domain gap''. It does not need heavy robotic effort. 

In experiments, we trained several YOLOv5x networks to understand the performance of the proposed method. The training data was collected using the proposed and several other methods. The results confirmed data collected using the proposed method do have claimed advantages. We also conducted multiple ablation studies to look into the impact of data volumes and ratios when training detection neural networks using data collected with the proposed method. The results provided a good guide on optimizing data configurations and parameter settings for training detectors.

The contributions of this work are as follows. (1) We develop an automatic data-collection method in which a robot holds target objects and observes them. The method yields observation images and target regions segmented from the images. (2) We develop a copy-paste image synthesis method to enrich the training data. The method pastes object regions on various rack backgrounds to balance ``domain randomization'' and ``domain gap''. The rack backgrounds are also automatically collected by the robot. (3) We examined combinations of the observation and synthetic images and compared them with other data sets to understand the impact of data volume and ratios.

The remaining part of this paper is organized as follows: Section \ref{sec:related_work} reviews related work. Section \ref{sec:sys_hardware_data_pre_workflow} presents the hardware system and the proposed method's workflow. Section \ref{sec:detail_implem} delivers technical details. Section \ref{sec:experiment} shows experiments and analysis. Section \ref{sec:conclusions} draws conclusions.

\section{Related Work}
\label{sec:related_work}

We review the related work considering robotic data collection and data synthesis, respectively.

\subsection{Automatic Data Collection Using Robots}


Segmenting the object regions from a picture is the basis of automatic data collection. Conventional methods used simple backgrounds \cite{sapp2008fast}, known environments \cite{suchi2019easylabel}, or designed easily identifiable gadgets \cite{kiyokawa2019fully}\cite{gregorio2020} to simplify object extraction. The methods required careful preparation about scenes and objects.

Robot-based methods leverage actuated robots to simplify object segmentation. They can be traced back to early studies in object recognition and 3D object modeling \cite{fitzpatrick2003grounding}\cite{welke2010autonomous}\cite{browatzki2012active}\cite{krainin2011manipulator}. These work took advantages of robotic manipulation sequence to perceive objects from different viewpoints and segment the objects from the background. From the robotic manipulation perspective, such segmentation can be divided into two categories: In-hand object segmentation and Interaction-based segmentation.

\subsubsection{In-hand object segmentation}
Previous work on in-hand object segmentation used known robot models and handcrafted visual features to isolate in-hand objects from background environments and robot hands. For example, Krainin et al. \cite{krainin2011manipulator} isolated in-hand objects' point clouds by examining the Euclidean distance to the robot model. Welke et al. \cite{welke2010autonomous} segmented in-hand objects from images based on Eigen background subtraction, disparity map, and hand localization. These methods required manually preparing detectors for various targets considering their visual features. 

More recent studies used deep learning to reduce the reliance on hand-crafted visual features for in-hand object segmentation. For instance, Florence et al. \cite{florence2020robot} proposed a self-supervised framework to segment in-hand objects. The framework involved two steps that used the same training and learning routine. In the first step, the authors generated masks for the robot by considering combined depth and RGB information, and trained a neural network model based on the masks to differentiate the robot from the background. In the second step, the authors masked the grasped object and train neural network models to isolate the object from the robot hand. Boerdijk et al. \cite{boerdijk2020self} used optical flow to respectively segment manipulators that were holding and not holding objects. The segmented data set were used to train a neural network for isolating manipulators and grasped objects.

\subsubsection{Robot-object interaction}
On the other hand, some researchers took advantages of non-prehensile robot manipulation like push to change object perspectives and segment them based on robot motion cues \cite{pathak2018learning}\cite{bjorkman2010active}\cite{faulhammer2016autonomous}. For example, Pathak et al. \cite{pathak2018learning} designed a framework to continuously refine a neural network model that generates object segmentation masks through robot interaction. The model initially generated hypothesis segmentation masks for objects. The masks were refined based on the pixel differences of the images captured before and after robotic interactions. The generating model was updated along with the refined masks. Singh et al. \cite{singh2021nudgeseg} proposed to segment unknown objects in a cluttered scene while repeatedly using robotic nudge motions to interact with objects and induce geometric constraints. Robotic interactive segmentation often requires a static scene or surface to permit interaction between robots and objects \cite{schiebener2014physical}\cite{eitel2019self}. It is more complicated compared with the in-hand object segmentation as the object poses needs to be controlled and changed through robotic manipulation.

A critical problem of the robotic methods is that they are unsuitable for preparing a large amount of training data as robots consume much time and energy to perform the physical motion. Conducting thousands of robotic motion trajectories to collect data is impractical. Also, the robots in the systems are fixed, have limited views, and can only collect data in a narrow range of scenarios. Neural networks trained using the collected data may suffer from contextual (background) bias and have bad generalizability  \cite{divvala2009an}\cite{barnea2019exploring}. 

This study focuses on robotic data collection while considering leveraging data synthesis to reduce robotic usage. We first ask the robot to hold a single tube and annotate the tube's bounding polyhedron by extracting in-hand point cloud according to the robot's tool center point (TCP) and hand model. Then, we map the annotated bounding polyhedron to 2D image regions in the robot's camera view for extracting the tube region. The robot moves the tube to different positions and rotations to obtain many varieties of 2D images and tube regions. The images and tube regions are respectively used for training and synthesizing new data in a later stage. 

\subsection{Data Augmentation and Synthesis}
Data augmentation and synthesis are the two most well-used methods to enrich training data. Data augmentation generates new data by transforming the existing training data with specific rules or learning-based methods. Data synthesis generates synthetic data by merging existing data with others or using computer simulations. Concurrent publications tend to mix these nomenclatures. Therefore we conduct a uniform literature review of them below without differentiation.


The copy-paste method is widely used for generating synthetic data. It segments foreground objects from existing images, possibly modifies them, and pastes them onto new backgrounds \cite{georgakis2017synthesizing}\cite{dwibedi2017cut}\cite{ghiasi2021simple}. The copy-paste method is easy to implement and shows notable performance over using pure real data. Previous studies showed that it was important to carefully select the backgrounds when pasting objects. For example, Divvala et al. \cite{divvala2009empirical} experimentally showed visual context benefited object detection performance and reduced detection errors. Dvornik et al. \cite{dvornik2018modeling} showed that the correct visual context when pasting object can improve prediction performance while inappropriate visual context led to negative results. Wang et al. \cite{wang2020constrained} swapped objects of the same class in different images to ensure contextual consistency between objects and backgrounds and showed using the exsiting backgrounds had better performance than random ones. Also, the copy-paste method requires a data set containing many possible views of the object that are easy to be cut out. It is burdensome for humans to prepare them. 

Graphical simulation is another popular method for synthesizing training data. The benefits of simulation is that it allows freely changing light conditions and materials to increase variation. It also allows capturing many views of objects by simply transforming virtual camera poses. For example, Hoda{\v{n}} et al. \cite{hodavn2019photorealistic} and Richter et al. \cite{richter2016playing} respectively used photo-realistic rendering to synthesize images of 3D object models and scenes. The methods required a lot of computational resources to narrow down the domain gap between synthetic and realistic data. Tobin et al. \cite{tobin2017domain} proposed the concept of domain randomization (DR). They randomized a simulator to expose models to a wide range of environments and obtain varied training data. Instead of photo-realistic rendering, the method only required low-fidelity rendering results to reach satisfying accuracy for medium-size objects. Carlson et al. \cite{carlson2018modeling}, Hinterstoisser et al. \cite{hinterstoisser2019an}, Prakash et al. \cite{prakash2019structured}, and Tremblay et al. \cite{tremblay2018training} respectively used DR to narrow down the domain gap. The authors randomly changed the context in simulation so that ``the real data was made to be just like another simulation'' \cite{tsirikoglou2020survey}. Yang et al. \cite{yang2022image} and Sundermeyer et al. \cite{sundermeyer2018implicit} respectively sampled viewpoints of 3D object models using simulation and mixed the samples with real backgrounds to reduce the human effort for preparing scenes with rich domain randomness. Besides DR, Generative Adversarial Networks (GANs) were also promising to reduce domain gap. For example, Chatterjee et al. \cite{chatterjee2022enhancement} designed a lightweight-GAN to synthesize data for training plastic bottle detectors. 

In this study, we leverage data synthesis to enrich the training data. We develop a copy-paste based method to attach tube cap regions separated from robotic observation images to rack backgrounds and thus synthesize new images. Various constraints like rack dimensions and tube occlusions can be considered during the synthesis to reduce the domain gap. The synthetic data is mixed with real-world data to promote the performance of YOLO-based tube recognition neural networks. It is also compared with other data collection methods to understand the influence of data volume and data combination ratio.



\section{Robot System and Workflow} 
\label{sec:sys_hardware_data_pre_workflow}
\subsection{Configurations of the Robot System}
\begin{figure}[!htbp]
    \centerline{\includegraphics[width=\linewidth]{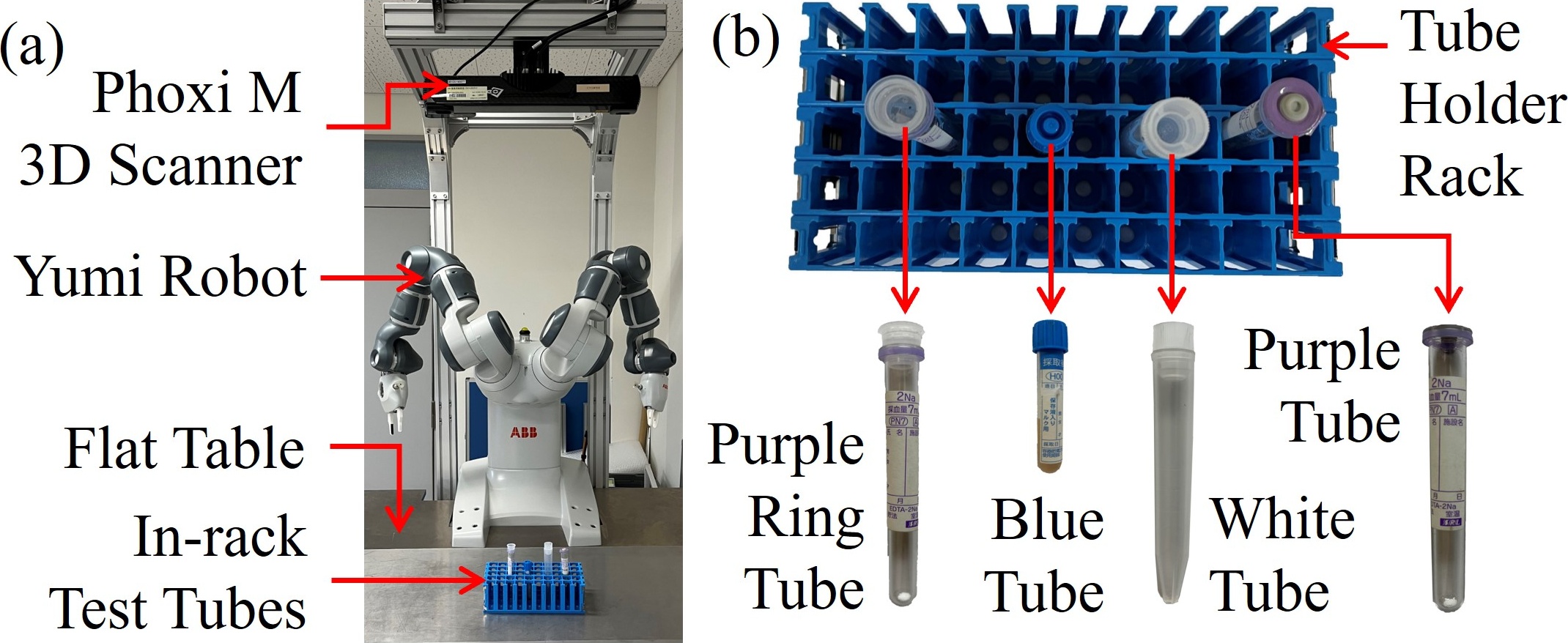}}
    \caption{(a) The system configuration. (b) The test tubes and rack in the view of the Phoxi M 3D Scanner.}
    \label{fig:env}
\end{figure}

\begin{figure*}[!htbp]
    \centerline{\includegraphics[width=\textwidth]{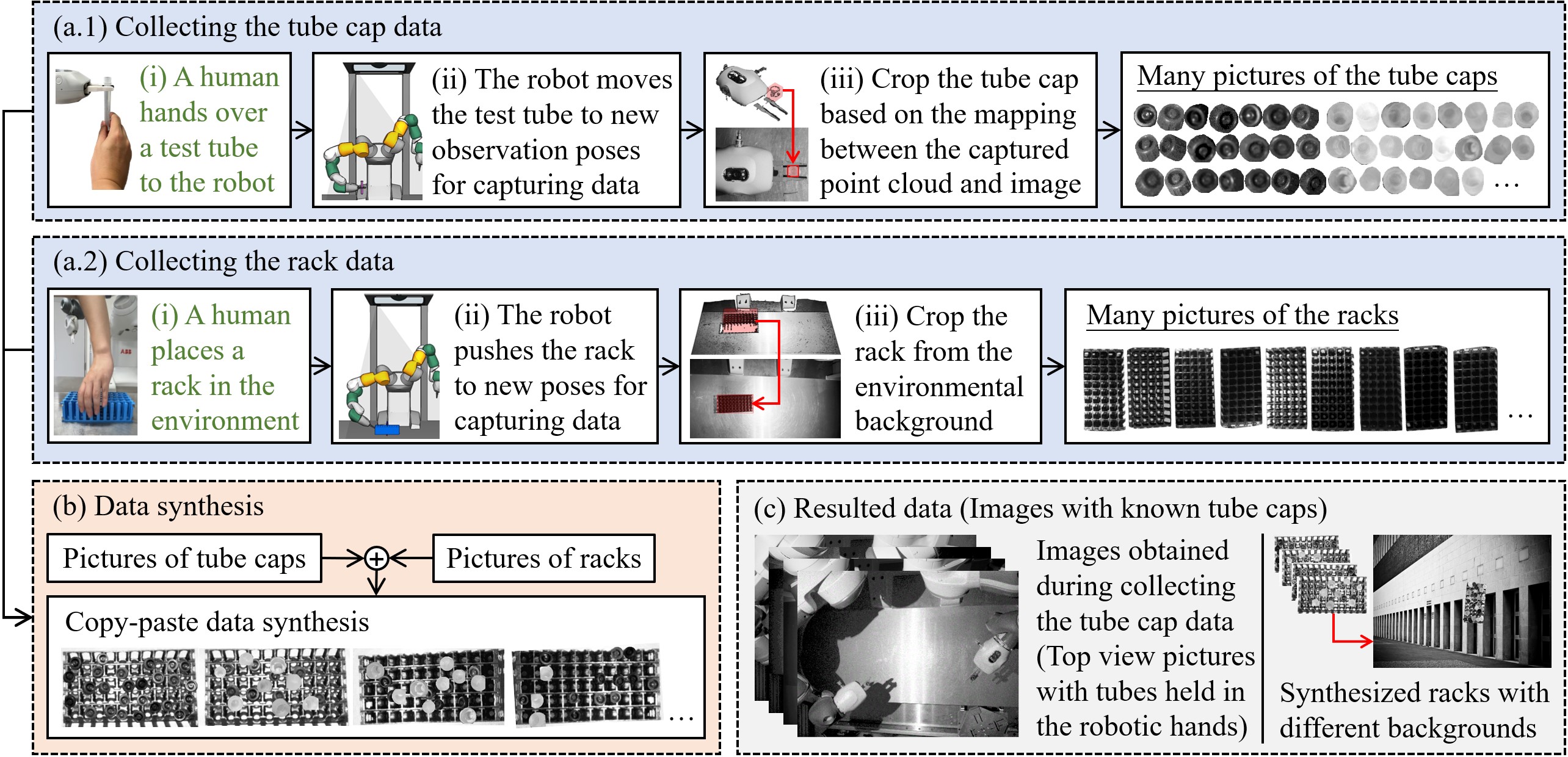}}
    \caption{Workflow of the proposed method. (a.1,2) Data collection component. (b) Data synthesis component. (c) Resulted data.}
    \label{fig:workflow}
\end{figure*}
Fig. \ref{fig:env}(a) shows our robot system used for preparing the training data. A Photoneo Phoxi M 3D Scanner is used for capturing objects on the flat table. An ABB Yumi dual-arm robot with a two-finger gripper is used to manipulate objects in the system. A flat table is set up in the front of the Yumi robot. The in-rack test tubes to be recognized are placed on the surface. The Phoxi scanner is a structured-light based depth sensor. It can capture gray images and point clouds simultaneously. Each data point of a point cloud captured by the Phoxi scanner have a one-to-one correspondence to a pixel in a gray image. We can segment an object in the gray image by considering its point cloud.


Especially, we install the Phoxi scanner on top of the robot to obtain a top view of the racks and tubes. When recognizing tubes in the rack, we select the tube caps as the primary identifiers. There are two reasons why we prefer using the tube caps for identification. The first one is that obtaining the point cloud of a translucent or crystal test tube fails easily due to limitations of the structured-light based depth sensors. The second one is that the tube bodies are blocked by the caps and also occluded by surrounding tubes when placed in the rack and viewed from a top position. They are less visible. However, despite the reasons and their merits, there is a problem that different types of tubes may share a same cap type. In this work, we assume the test tubes with the same caps can be identified by their heights in the rack and analyze the point cloud to differentiate them. 


\subsection{Workflow for Data Preparation} 
We prepare the training data using the robot system following the workflow shown in Fig. \ref{fig:workflow}. There are four dashed boxes in the chart, where (a.1) and (a.2) have a blue background color and represent the data collection component, (b) has an orange background color and represents the data synthesis component, (c) has a gray background and represents the resulted data.

The first blue dashed box (Fig. \ref{fig:workflow}(a.1)) comprises three steps. First, a human hands over an unknown test tube to the robot. The tube is assumed to be grasped vertically by the robot after handover, with the tube cap left above the robotic fingertips. Second, the robot moves the test tube to the observation poses prepared offline while considering avoiding self-occlusions. The Phoxi sensor will capture the test tube's gray image and point cloud at each observation pose. Third, the system segments the cap region out of the captured image based on a mapping from its counterpart point cloud. The segmentation result only includes the cap. The background will be removed thanks to the point cloud mapping. The output of this dashed box includes many cap region pictures. They are observed from different views and thus have different illumination and visual conditions.

The second blue dashed box (Fig. \ref{fig:workflow}(a.2)) is similar to the first one and also comprises three steps. First, a person places a rack in the environment. Then, the robot pushes the rack to random poses, capturing the rack's gray image and point cloud at each pose. Third, the system segments the rack region out of the captured image based on the mapping from the rack's counterpart point cloud. The result of this dashed box includes many rack region pictures. Like the caps, the rack region pictures also have different illumination and visual conditions since the data is captured from different view positions. 

The orange dashed box shows the data synthesis process, where the cap region pictures obtained in the first ``Data Collection'' dashed box are pasted onto the rack region pictures obtained in the second ``Data Collection'' dashed box for synthesizing new images. Constraints like rack boundaries and overlapping caused by perspective projection are considered during the synthesis. The output of the dashed box will be racks filled with many tube caps. The ``Copy-paste data synthesis'' sub-block illustrates several examples of the output. 

The final data preparation results include the images obtained during collecting the tube cap data (observation images) and the synthetic images. They are illustrated in the gray dashed box (Fig. \ref{fig:workflow}(c)).

Note that the above workflow is not completely automatic. The sub-blocks with texts highlighted in a green color involve human intervention. Also, before data collection, we need to prepare the camera calibration matrix and test tube observation poses. The camera calibration matrix transforms the point cloud captured in the camera's local coordinate system into the robot coordinate system. Many existing methods exist for obtaining the calibration matrix \cite{lundberg2015intrinsic}. To avoid repetition, we don't discuss the details in this manuscript. The test tube observation poses are a set of tube positions and rotations for the robot to hold and capture observation images. The developed method will generate robot joint configurations considering the robot grasping and tube observation poses. Section \ref{sec:detail_implem} will present detailed algorithms on the generation.

\section{Implementation Details}
\label{sec:detail_implem}
 
\subsection{Observation Poses for Collecting Tube Caps}
When collecting the tube cap data, the robot moves the tube held in its hand to different poses for observation. The observation poses are generated considering two constraints: (1) Diversity of the captured cap data; (2) Occlusions by robot links. Taking into account these two constraints allow us to include the tube caps from many viewpoints and thus cover lots of illumination and visual conditions. Meanwhile, they help to prevent the robot links from occluding the grasped test tubes and make sure the tubes are visible to the vision sensor.

\begin{figure}[!htbp] 
    \centerline{\includegraphics[width=\linewidth]{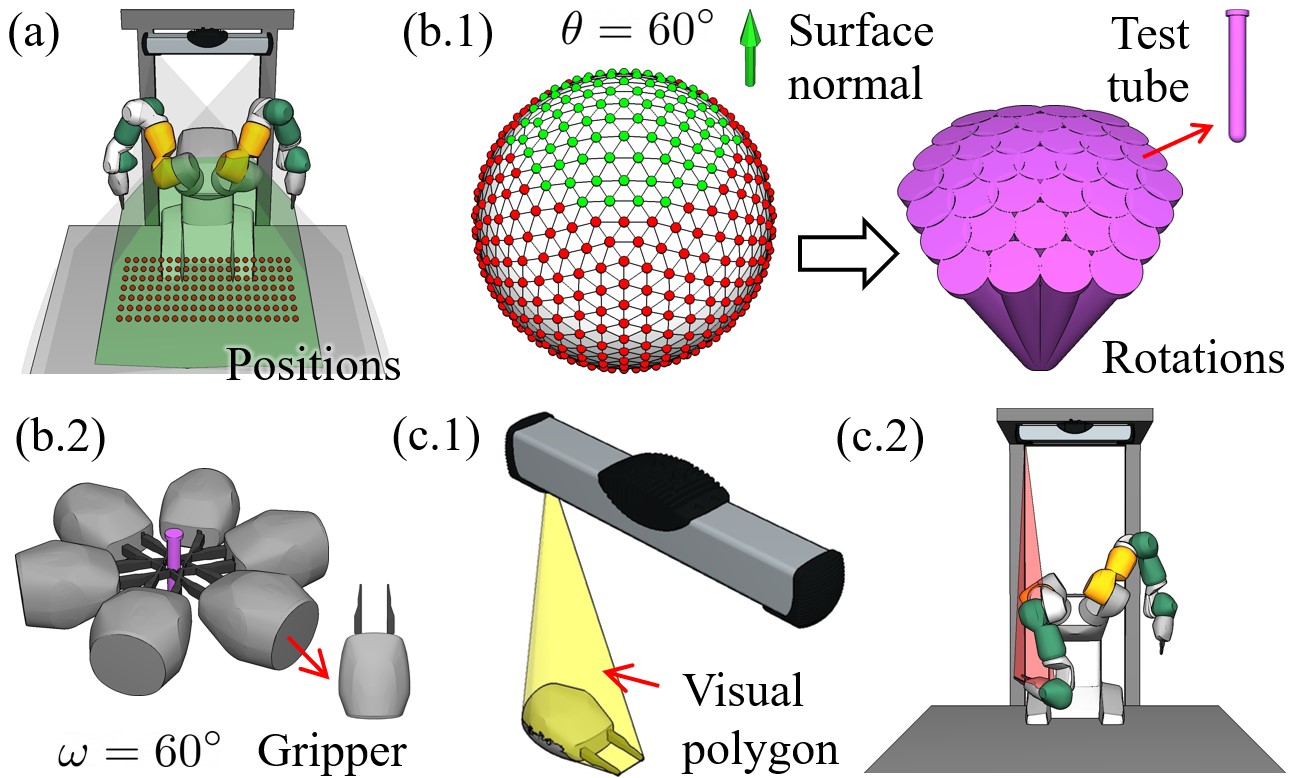}}
    \caption{(a) Sampling observation positions. The green region is the visible area of the Phoxi scanner. The red points are the sampled positions. (b.1) Sampling rotations based on a level-four icosphere. The left spherical polyhedron illustrates the icosphere. The green vertices are the ends of feasible vectors that have less than $\theta=60^{\circ}$ angles with the surface normal. They imply the tube rotation poses shown on the right. (b.2) The grasping poses for each sampled tube pose form a SO(2) group. They are sampled considering an interval $\omega$ for numerical analysis. (c.1) A visual polyhedron computed using the camera origin and vertices of the robot hand model. (c.2)  The grasped object has a risk of being occluded by the robot arm when there is a collision between the visual polyhedron and the robot arm links.}
    \label{fig:offline_obs_pose}
\end{figure}

Fig. \ref{fig:offline_obs_pose} illustrates the observation pose generation process and how the two constraints are taken into account in it. First, we sample the positions and rotations of a tube held by the robot hand uniformly in the Phoxi depth sensor's visible range. Tube data captured under the sampled poses will have rich light conditions and a large variety of visible tube edges for training a recognition neural network. Especially, the tube rotations are sampled according to the vertices of a level-four icosphere \cite{wan16}. An icosphere is a spherical polyhedron with regularly distributed vertices. The vectors pointing to the vertices of an icosphere help to define the rotations of a tube\footnote{A tube is centeral symmetric. We do not need to consider its rotation around the central axis. The vectors pointing to the vertices of an icosphere can thus define a tube pose.}. A level-four icosphere has $642$ vertices and thus leads to $642$ vectors and test tube rotation poses. Thanks to the visibility constraints, we do not move a test tube to all of the rotation poses for capturing data as the tube caps facing downward will not be seen by the Phoxi sensor. We filter the $642$ vectors by considering their angles with the normal of the table surface for placing a rack. The vectors with large angles from the surface normal cannot be seen and will not be considered. The spherical polyhedron in Fig. \ref{fig:offline_obs_pose}(b.1) illustrates the level-four icosphere. Vectors pointing to the red vertices have more than $\theta$ angles from the surface normal and are removed. The green vertices are the remaining candidates. The purple tube bouquet on the right side of Fig. \ref{fig:offline_obs_pose}(b.1) illustrate the tube poses implied by vectors pointing to the remaining candidate vertices.




Next, we plan the robot motion to move the test tube held in a robot hand to the sampled tube positions and rotations. We assume a test tube is vertically grasped at the finger center of a robot hand. Since a tube is central symmetric, many grasping poses meet the assumption. The grasping hand may rotate freely around the symmetry axis of the test tube, as shown in Fig. \ref{fig:offline_obs_pose}(b.2). The rotation is compact and forms a SO(2) group. For numerical analysis, we sample the rotation in the SO(2) group with a rotation interval hyperparameter named $\omega$ to obtain a series of discretized grasping poses. The hand illustrations in Fig. \ref{fig:offline_obs_pose}(b.2) are the grasping poses obtained with $\omega$ = 60$^\circ$. The sampled grasping poses provide many candidate goals for robot motion planning and thus increase the chances of successfully moving and observing the tube.

When determining which exact candidate goal to move to, we examine the occlusions from the robot arm links and avoid choosing the grasping poses that lead to invisible tubes. In detail, examining the occlusion is done by checking the collision between a visual polyhedron and the robot arm links. The visual polyhedron is computed using the camera origin and vertices of the robot hand model, as illustrated in Fig. \ref{fig:offline_obs_pose}(c.1). The robot arm may occlude the tube and the vision sensor fails to capture it when there is collision between the visual polyhedron and the robot arm links. Fig. \ref{fig:offline_obs_pose}(c.2) exemplifies such a case.

\subsection{Using Annotation Masks to Segment Cap Pictures}
\begin{figure}[!tbp] 
    \centerline{\includegraphics[width=\linewidth]{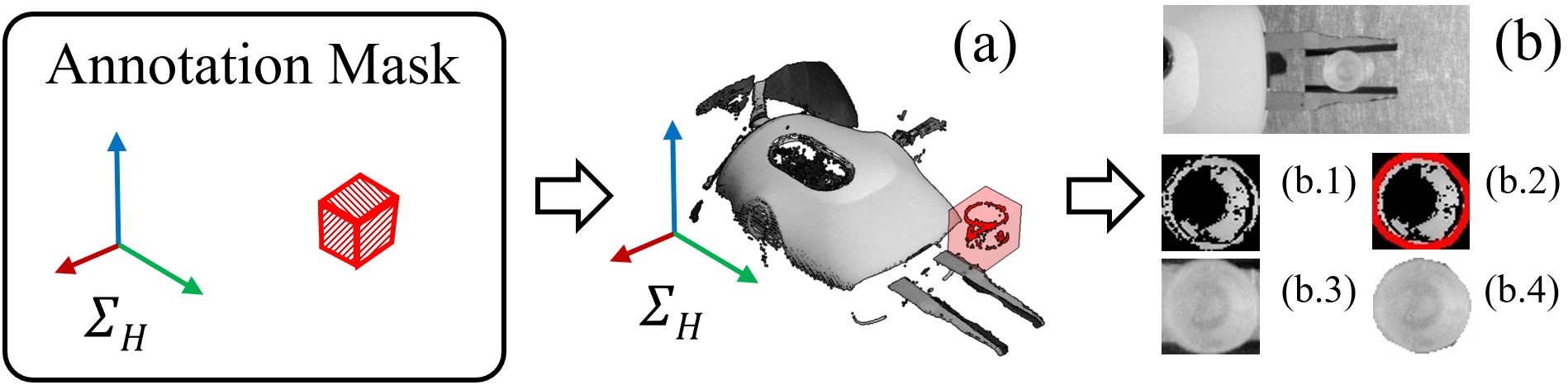}}
    \caption{Workflow for extracting the cap picture using an annotation mask. (a) Applying a mask described in the local coordinate system of the holding robot hand to the captured point cloud. (b) The extract point cloud is projected back to the 2D grey image for segmenting a picture of the cap region. (b.1) The back-projected results might be disconnected pixels. (b.2) A bounding convex hull of the disconnected pixels is computed. (b.3,4) The cap region is segmented based on the bounding convex hull.}
    \label{fig:ann_mask}
\end{figure}

Since the tube is handed over from a human and the Phoxi sensor captures the cap data from many different views, the captured tube point clouds change dynamically and have noises. It is unstable to extract cap point clouds by autonomously detecting them. Thus, instead of autonomous detection, we prepare an annotation mask in the robot hand's local coordinate system to help extract the test tube cap's point clouds. The extracted point clouds will be back-projected to the corresponding 2D grey image for segmenting a picture of the cap region. Fig. \ref{fig:ann_mask} shows the details of this mask and how it helps to segment the cap regions. The mask and back projection enable us to precisely segment the cap regions while avoiding including backgrounds.

\begin{figure}[!tbp]
    \centerline{\includegraphics[width=\linewidth]{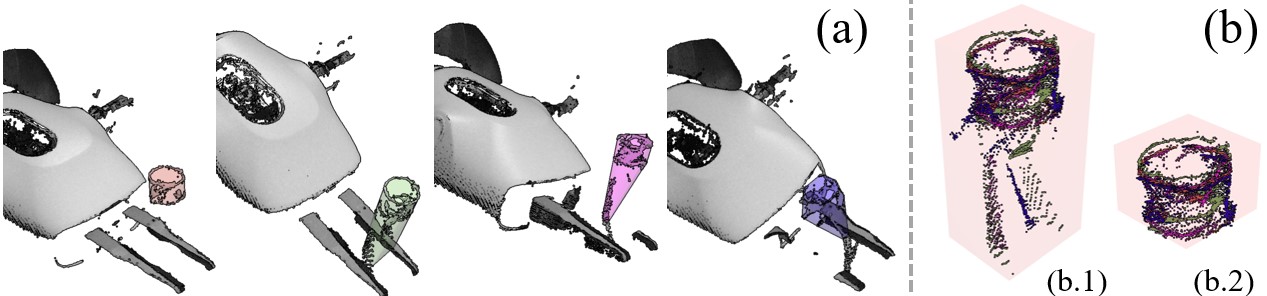}}
    \caption{(a) Capture data from different views. The tube cap's point clouds are obtained by examining the area on top of the holding fingers. They are high lighted with colored polyhedrons. (b) Merge the cap's point clouds in (a) under the robot hand's local coordinate system, and compute a bounding box of the merged result as an annotation mask. (b.1) Raw bounding box. (b.2) The bounding box can be adjusted interactively if needed.}
    \label{fig:mask_extract}
\end{figure}

To prepare an annotation mask, we move the robot hand that holds a test tube to a fixed position under the Phoxi sensor and trigger the sensor to capture a point cloud. We can easily get the cap's point cloud data by examining the area on top of the holding fingers and obtain an annotation mask by considering a bounding polyhedron of the data. However, a single bounding polyhedron may not be general for others since the captured point cloud is susceptible to light reflection or perspective projection (self-occlusion). Thus, instead of a single point cloud and polyhedron, we collect point clouds from multiple views, merge them under the robot hand's local coordinate system, and compute a bounding box of the merged result as an annotation mask. Fig. \ref{fig:mask_extract} shows an example. The multiple views are sampled the same way as the observation poses mentioned in the previous subsection. However, we do not need to change the observation positions since we aim to obtain a bounding box mask in the hand's coordinate system. The views under various rotations could provide enough superficial point cloud data to meet the requirements. Note that the merged result may include noise point data induced by reflections from the transparent tube body and lead to a mask larger than the cap. We provide an interactive user interface for manually adjusting the bounding box sizes and minimizing the negative influences caused by the noises. The adjustment is optional and may be performed when precisely segmenting the cap region is demanded.

\subsection{Copy-Paste Synthesis}
We apply random scaling, blurring, brightness, and contrast to the segmented tube caps and then paste them onto the segmented rack background for data synthesis. During pasting, we permit the overlap among the cap regions to approximate tube-to-tube occlusion. After pasting, we randomize the environmental background (background of the rack) to narrow further the domain gap between synthetic images and images captured in the real world.

A critical maneuver here is that we consider the co-occurrence of the test tubes and the rack and paste the tube cap pictures onto a rack instead of random backgrounds like \cite{boerdijk2021s}. We randomly sample positions inside rack pictures for pasting tube caps and use a pasting number $T$ to control the clutter. Note that there is no need to exactly paste a tube cap near the hole centers of a rack as the tubes tilt randomly inside the rack holes. The visible cap regions may reasonably overlap with a hole boundary or other holes.

\begin{figure}[!htbp]
    \centerline{\includegraphics[width=\linewidth]{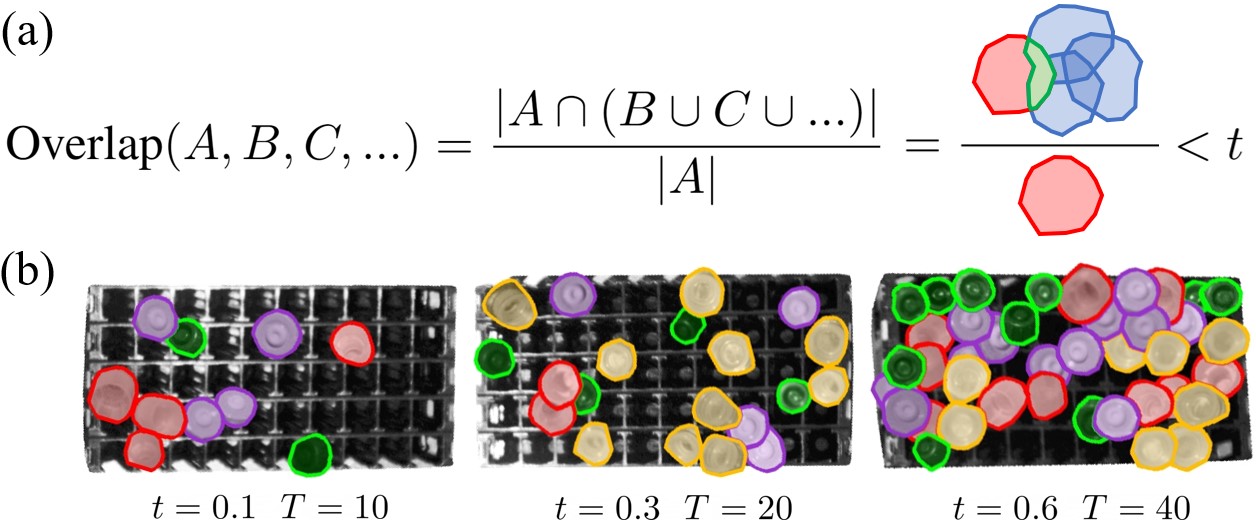}}
    \caption{(a) Using a threshold $t$ to simulate cap occlusions. ``A'' represents a previously pasted cap region. ``B'', ``C'', ... represent the caps pasted after ``A''. (b) Results with different $t$ values.} 
    \label{fig:overlap_constraints}
\end{figure}

For tube-to-tube occlusion, we consider the perspective projection of a vision sensor and define an occlusion threshold $t$ to permit overlap among the visible cap regions. A vision sensor's perspective projection leads to mutual occlusions in the rack at certain viewpoints. The occlusion threshold helps to simulate the occlusion and defines the maximum percentage that segmented cap pictures can overlap or occlude. Fig. \ref{fig:overlap_constraints} shows how the $t$ threshold works. It adds a constraint to pasting, where a previously pasted cap picture ``A'' must have less than $t$ percentage overlap with the union of caps pasted later. The $B\cup C \cup...$ component in the nominator of Fig. \ref{fig:overlap_constraints} implies the union of caps pasted after ``A''. When a new cap is randomized, it must be unioned with this component to ensure the $t$ constraint on all previous ``A'' is not violated. There are two noticeable points for $t$. First, its value could be devised respectively considering the heights of specific tube types. Second, its value is correlated with the pasting number $T$. The maximum number of pasted tube caps in a rack that meet the $t$ threshold may be less than a given $T$. In that case, we constrain the maximum number of pasted tube caps to the smaller value to ensure $t$ is not invalidated.


For the environmental background, we use the BG-20k data set \cite{li2022bridging} to obtain high-resolution random background images and change the background of a synthetic image with a $0.5$ probability.

\section{Experiments and Analysis}
\label{sec:experiment}

\begin{figure}[!htbp]
    \includegraphics[width=\linewidth]{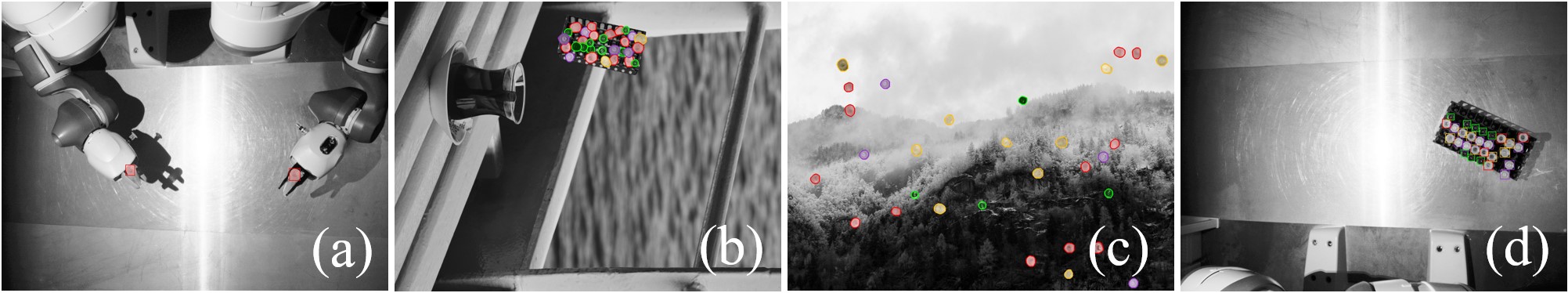}
    \caption{Exemplary images collected using the various methods. (a) RO. (b) SR. (c) SB (d) CL.} 
    \label{fig:alldata}
\end{figure}

We carried out experiments to compare YOLOv5x \cite{glenn_jocher_2020_4154370} detectors trained using data sets collected with the proposed method and several other methods to understand the performance. Table \ref{tab:evalution_results} shows the methods. The SR (Synthesis by pasting to Racks) method pastes randomly selected cap pictures onto rack backgrounds to synthesize training data. It represents the synthesizing method used in this work. The SB (Synthesis by pasting to BG-20k) method is an alternative synthesis method. Instead of being pasted onto a rack, randomly selected cap pictures are pasted to random backgrounds selected from the BG-20k data set. The RO (Robotic Observation) method is a byproduct of robotic cap segmentation, where the robot holds test tubes for data collection. We considered RO an independent method because we wondered if the hand-held observation was enough for training. We also combined RO, SR, and SB methods (the ** row) to see if they help achieve a satisfying performance. The RO+SR combination is exactly our proposed method in this work. We especially proposed it since RO is a pre-process of robotic cap segmentation. Using combined RO+SR does not increase effort. Combining RO+SB or RO+SR+SB are also candidate choices. They have the same cost as using independent SR or SB data\footnote{Synthesizing data is considered to be free as it only require computational work. Thus, the costs of SR and SB depend on the RO process.}. Finally, the CL (Crowd-source Labeling) method is a conventional one that requires humans to place racks with tubes under the robot and label the captured images manually. Fig. \ref{fig:alldata} shows exemplary images collected using the different methods. 

\begin{table}[!htbp]
\caption{Summary of the data collection methods}
\label{tab:evalution_results}
\centering 
\begin{tabular}{lll}
\toprule
Abbr. & Full Name & Description\\
\midrule
SR & Synthesis by pasting to Racks & Caps on racks\\
SB & Synthesis by pasting to BG-20k & Caps on random background\\
RO & Robotic Observation & Tubes held in robotic hands\\
** & Combinations of above methods & SR+SB is the proposed one\\
CL & Crowd-source Labeling & Tubes in a rack on the table\\
\bottomrule
\end{tabular}
\end{table}

\subsection{Performance of Various Data sets} 

We collected various data sets with the methods and their combinations, used the data sets to train YOLOv5x detectors, and examined the performance of the trained detectors using a testing data set for comparison.

The first data set is CL${}^{200}$. It is considered a baseline for comparison. In collecting the data set, we collected $200$ images with random tube and rack states and labeled the tube regions manually using LabelImg\footnote{https://github.com/heartexlabs/labelImg}. There are, in total, $5916$ labeled instances in the $200$ images.

\begin{figure}[!htbp]
    \includegraphics[width=\linewidth]{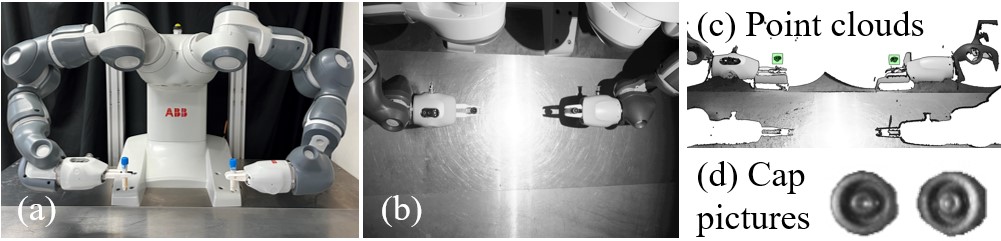}
    \caption{(a) The robot moves test tubes for observation. Both arms are used. (b) Observation Image. (c) Point clouds captured by the Phoxi sensor. (d) Cap pictures segmented from the observation image.} 
    \label{fig:collectrob}
\end{figure}

The second data set is SR${}^{1600}$. In order to collect it, we first prepared many cap pictures using robotic observation. As shown in Fig. \ref{fig:env}(b), we assumed four different test tubes and took advantage of the Yumi robot's both arms to collect cap data quickly. For each tube type, we handed over two same ones to the two robotic arms for observation. Each arm moved its held tube to $400$ observation poses for data collection. See Fig. \ref{fig:collectrob}(a) for example. Here, we set the hyperparameter $\theta$ and $\omega$ to $30^{\circ}$ and $360^{\circ}$ (single grasping pose) and set the positions to be evenly sampled on the table with a granularity of $0.1m$ for generating the observation poses. In total, more than $400$ observation poses were obtained under the parameter setting for each arm, and we used the first $400$ for collecting images. As a result, we obtained $400$ observation images ($800$ cap pictures since there are two tubes in each image, see Fig. \ref{fig:collectrob}(b) for example) for a single tube type and $1600$ observation images for all tube types. We segmented $3200$ pictures of cap regions from the observation images considering point cloud mapping. Fig. \ref{fig:collectrob}(c) shows the collected point clouds with highlighted caps (green). Fig. \ref{fig:collectrob}(d) shows the segmented cap regions. Besides the cap regions, we collected $15$ images with racks (a single rack in each image) and segmented $15$ pictures of racks. We synthesized a data set of $1600$ images by pasting caps randomly selected from the $3200$ cap pictures to racks randomly selected from the $15$ rack pictures (SR method). During synthesis, we set the pasting number to be $T = 30$, and set the occlusion threshold for the ``Blue Tube'' to be $t_{blue} = 0.4$ and other tubes to be $t_{others} = 0.15$. We chose these parameter settings because the ``Blue Tube'' was shorter and susceptible to occlusion. We increased its occlusion threshold to mimic frequent visual blockage from other tubes. Also, we increased the variety of the segmented cap pictures by applying random scaling ($0.9\sim1.1$ of original picture size, $0.5$ probability), random blur ($3\times3$ kernel, $0.5$ probability), random brightness ($0.9\sim1.1$ of original brightness, $0.5$ probability), and random contrast ($0.9\sim1.1$ of the original contrast value, $0.5$ probability) using the Albumentations\footnote{https://albumentations.ai/} library. The background of the rack was randomly chosen from the BG-20k data set with a $0.5$ changing probability.

The third data set is RO${}^{1600}$. It is a semi product of robotic cap segmentation and comprises the $1600$ observation images obtained during robotic observation.

The fourth data set is SB${}^{1600}$. In contrast with the SR${}^{1600}$ data set, we pasted randomly selected caps directly to images from the BG-20k data set for obtaining data. The pasted caps might freely distribute on the image background. The segmented racks were not used. The pasting number $T$ and occlusion threshold $t$ are $35$ and $0.15$ respectively. There was no difference on $t$ for different tubes. The randomization were performed in the same way as obtaining SR${}^{1600}$.

We also used combined methods to collect data sets and study if the combination led to better results. The combined data sets include RO${}^{1600}$+SR${}^{800}$, RO${}^{1600}$+SB${}^{800}$, RO${}^{1600}$+SR${}^{400}$+SB${}^{400}$, SR${}^{800}$+SB${}^{800}$. Here, the superscript number on the upper-right of a method name means the number of images collected using the method. The ``$+$'' symbol indicates that the data sets comprise data collected using different methods. The RO${}^{1600}$+SR${}^{800}$ data set represents the data collected using the proposed method. 

The left part of Table \ref{tab:evalution_results} summarizes the various data sets. They are used to train YOLOv5x detectors for comparison. Before training, the YOLOv5x detectors for all data sets were initialized with weights pre-trained using the COCO data set. The images in all data sets were regulated into a resolution of $1376\times1376$. Each data set is divided into a training subset and a validation subset according to a $4$ to $1$ data ratio. During learning, the training subset was fed to the training program with a batch size of $2$, and the training program performed validation per episode. The training process was stopped when the mAPs (mean Average Precision) \cite{zou2019object} for all objects reached higher than $99.0$\% under a $0.5$ IoU (Intersection over Union). Here, we defined a detected bounding box to be correct when its IoU with a ground truth cap bounding box was larger than $0.5$.

For evaluating the performance of YOLOv5x detectors trained using the various data sets, we collected a testing data set with $100$ images and labeled their ground truth using the same method as CL. We used the trained detectors to detect tubes in the testing data set. Like validation, we defined a detected bounding box as correct when its IoU with a ground truth cap bounding box is larger than $0.5$. We used the AP (Average Precision) metric to measure the detection performance of a single object class and used the mAP for all objects. Since the detector that met a single satisfying validation was not necessarily the best, we trained each detector twice and took the higher precision value on the testing data set as the final evaluation result.

\begin{table*}[htbp]
\renewcommand\arraystretch{1}
\caption{Comparison of detectors trained using different data sets}
\label{tab:evalution_results}
\centering 
\begin{threeparttable}
\begin{tabular}{llll|cccc|c}
\toprule
& & & & \multicolumn{4}{c|}{AP} &\\
\cmidrule(lr){5-8}
ID & Data Set Names & \# Caps & Remark & Blue & Purple & White &  Purple Ring  &  mAP \\
\midrule
1 & CL${}^{200}$ & $5916$ & Multiple tubes / image & $\mathbf{0.993}$ & $\mathbf{0.995}$ & $\mathbf{0.989}$  & $0.984$  & $\mathbf{0.990}$\\
2 & RO${}^{1600}$ & $3200$ & Two tubes / image & $0.380$ & $0.923$ & $0.695$  &  $0.630$  &  $0.657$\\ 
3 & SR${}^{1600}$ & $40000$ & $t_{blue}{=}0.4$ \& $t_{others}{=}0.15$, $T=25$ & $0.955$ & $ 0.979 $ & $0.871$  & $0.953$  & $ 0.940$\\
4 & SB${}^{1600}$ & $56000$ & $t{=}0.15$ (same for all tubes), $T=35$ & $0.808$ & $0.978$ & $0.812$  & $0.897$  & $0.874$\\
5 & RO${}^{1600}$+SR${}^{800}$ & $23200$ & See note 2 & $0.968$ & $\mathbf{0.995}$ & $0.978$ & $0.992$ & $0.983$ \\ 
6 & RO${}^{1600}$+SB${}^{800}$ & $31200$ & See note 2 & $0.881$ & $0.994$ & $0.971$  & $0.992$  & $0.959$\\ 
7 & RO${}^{1600}$+SR${}^{400}$+SB${}^{400}$ & $27200$ & See note 2 & $0.969$ & $0.994$ & $0.986$  & $\mathbf{0.993}$  & $0.986$\\ 
8 & SR${}^{800}$+SB${}^{800}$ & $48000$ & See note 2 & $0.973$ & $0.993$ & $0.969$  & $0.985$  & $0.980$\\
\bottomrule
\end{tabular}
  \begin{tablenotes}
  \item[Note 1:] Largest AP and mAP values are highlighted in bold.
  \item[Note 2:] The combined data sets are collected using the same parameters as respective ones.
  \end{tablenotes}
\end{threeparttable}
\end{table*}

Table \ref{tab:evalution_results} shows the evaluation results. We obtained the following observations and speculations from them.
\begin{enumerate}[noitemsep,topsep=0pt]
    \item[i)] Using the data set collected by robotic observation for training exhibited the worst performance, as shown by the 2nd row (RO${}^{1600}$).\\
    \underline{Speculation:} All images in the data set had a similar robotic background. They suffered from a domain shift.
    \item[ii)] The synthetic data sets do not necessarily lead to a good AP, as shown by the 3rd (SR${}^{1600}$) and 4th (SB${}^{1600}$) rows. The SR${}^{1600}$ data set exhibited higher performance than the SB${}^{1600}$ data set.\\
    \underline{Speculation:} The copy-paste synthesis failed to cover certain visual contexts; Pasting onto racks (SR) provided more effective visual contexts and benefited the neural network more than pasting onto random backgrounds (SB).
    \item[iii)] Combining the synthetic data sets with robotic observations is effective. It can be concluded by comparing the 5th, 6th, and 7th rows (RO${}^{1600}$+SR${}^{800}$, RO${}^{1600}$+SB${}^{800}$, RO${}^{1600}$+SR${}^{400}$+SB${}^{400}$) with the 2nd, 3rd, and 4th rows (RO${}^{1600}$, SR${}^{1600}$, and SB${}^{1600}$). The former rows had higher mAP than the latter.\\
    \underline{Speculation:} The robotic observation data set additionally provided helpful visual contexts.
    \item[iv)] The 5th row (RO${}^{1600}$+SR${}^{800}$) had a $2.4\%$ higher mAP than the 6th row (RO${}^{1600}$+SB${}^{800}$). Especially, the AP of the ``Blue Tube'' on the 5th row was $8.7\%$ higher than that on the 6th row. The AP of other tubes also had $0.1\%\sim0.7\%$ performance increase.\\
    \underline{Speculation:} Considering the rack as a local context helped improve domain-specific performance; The short ``Blue Tube'' could be easily blocked. The data set collected using the SR method had more simulated occlusions. They were important for recognizing the short ``Blue Tube''.
    \item[v)] The 7th row (RO${}^{1600}$+SR${}^{400}$+SB${}^{400}$) exhibited slightly higher mAP ($0.3\%$) than the 5th row (RO${}^{1600}$+SR${}^{800}$).\\
    \underline{Speculation:} Pasting onto racks (SR) provided better domain-specific features. Random backgrounds for the tubes slightly benefited the neural network and were less necessary if the goal context was limited.
    \item[vi)] The 5th row (RO${}^{1600}$+SR${}^{800}$) is competitive compared with the 1st row (CL${}^{200}$). The mAP was $0.7\%$ lower. The AP of the ``Blue Tube'' and ``White Tube'' were $2.5\%$ and $1.1\%$ lower, respectively. The AP of the ``Purple Tube'' was the same. The AP of the ``Purple Ring'' tube was $0.8\%$ higher.\\
    \underline{Speculation:} The robotic observation and paste-to-rack synthesis compensated for each other's shortcomings; There remained extreme cases that could be labeled manually but failed to be covered by robotic observation or synthesis, especially for the ``Blue Tube''.
    \end{enumerate}

\begin{figure}[!htbp]
    \includegraphics[width=\linewidth]{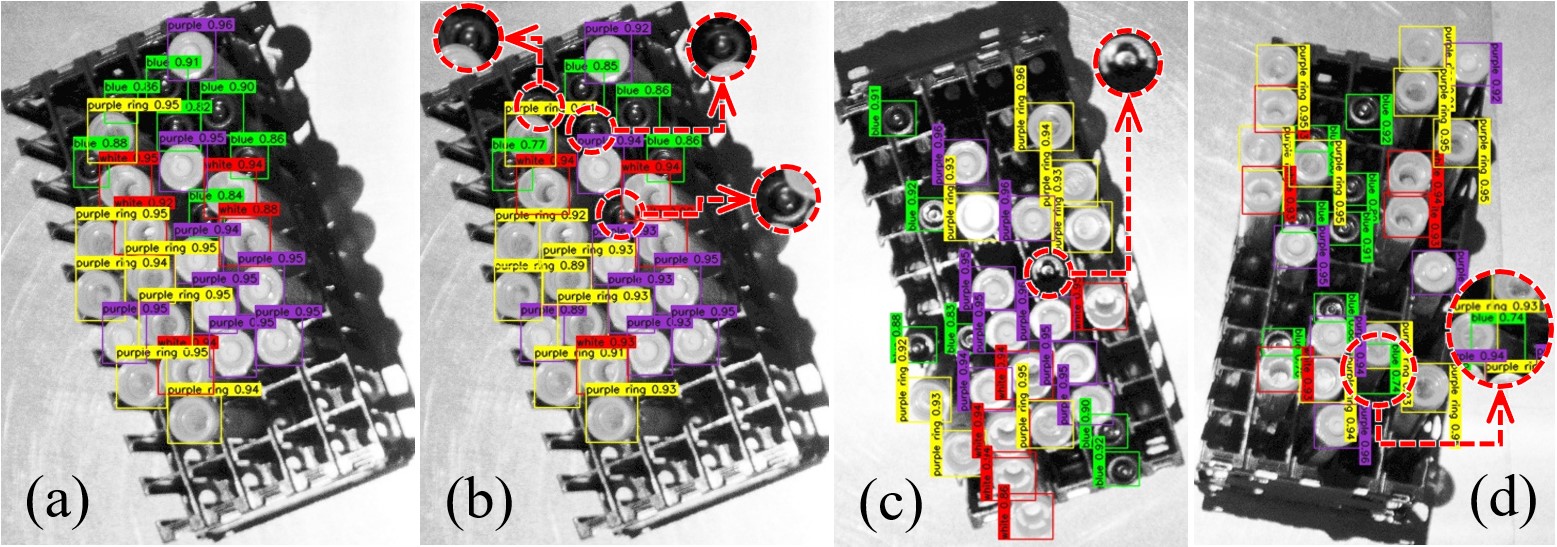}
    \caption{(a) Detector trained using the 5th data set successfully recognized all tubes. (b) Detector trained using the 6th data set failed to recognize the occluded tube in the red circle. (c) Detector trained using the 5th data set failed to recognize the shadowed tube in the red circle. (d) Detector trained using the 5th data set misrecognized the tube body in the red circle as a ``Blue Tube''.} 
    \label{fig:cases}
\end{figure}

Several failure cases are visualized in Fig. \ref{fig:cases} to provide the readers an insight into our observations and speculations. Fig. \ref{fig:cases}(a) and (b) exemplify the recognition results of detectors trained using the 5th (RO${}^{1600}$+SR${}^{800}$) and 6th data sets (RO${}^{1600}$+SR${}^{800}$). The latter one failed to recognize occluded tubes as the training data set had fewer simulated occlusions. The example is consistent with the observation and speculation in iv). Fig. \ref{fig:cases}(c) and (d) exemplify cases that the detectors trained using the 5th (RO${}^{1600}$+SB${}^{800}$) data set failed. In the first case, shadows from other test tubes were cast on a blue test tube cap. The detector failed to recognize the tube. In the second case, the detector misrecognized a crystal tube body as the ``Blue Tube'' cap due to the illusion caused by body-and-rack overlap. The two failure examples are consistent with the observation and speculation in vi). The synthetic data sets do not involve shadows or tube bodies. The detectors trained using them had worse performance in these cases than the one trained using the crowd-sourced real-world data.

In summary, the results of the various training data sets showed that combining data collected using the RO and SR methods was effective. The conclusion was satisfying as the RO method is a subset of the SR method. The workflow for collecting them is simple and clean. However, we wonder if the number of images in the RO data set could be reduced, as it needs much manual handover to collect them. This query prompted us to carry out the studies in the following subsection.

\subsection{Ablation Study} 
In this subsection, we conduct multiple ablation studies on the combined RO+SR data set to further understand 1) the influence of the data combination ratio and 2) the influence of pasting number $T$ and occlusion threshold $t$ used for generating synthetic data.  

\subsubsection{Influence of data combination ratio}
\label{subsec:blended_img_num}
The experiments for studying the influence of data combination ratio are divided into two parts. In the first part, we set the number of images collected using the RO method to $800$ and varied the number of images collected using the SR method from $200$ to $1600$ in a $2$-fold ratio to understand the importance of the SR data. The upper section of Table \ref{tab:number_real_syn} shows the precision of detectors trained using the varied data. The results indicate that the mAP improved when the SR image numbers increased from $200$ to $1600$. 
The second part is similar to the first one. In this part, we fixed the number of images collected using the SR method to $800$ and varied the number of images collected using the RO method from $200$ to $1600$ in a two-fold ratio to understand the importance of the RO data. 
The lower section of Table \ref{tab:number_real_syn} shows the precision of detectors trained using the varied data. The result indicates that the mAP improved when the RO image numbers increased from $200$ to $1600$. 

\begin{table}[htbp]
\caption{Influence of data combination ratio} 
\label{tab:number_real_syn}
\centering 
\begin{threeparttable}
\begin{tabular}{l|cccc|c}
\toprule
& \multicolumn{4}{c|}{AP} &\\
\cmidrule{2-5}
Data Set Names & Blue &  Purple & White &  Purple Ring  &  mAP \\
\midrule
RO$^{800}$+SR$^{200}$ & $0.958$  &  $0.994$ &  $0.973$    &  $0.987$   & $0.978$\\ 
RO$^{800}$+SR$^{400}$ & $0.964$  &  $0.992$ & $0.975$   &  $0.985$   &  $0.979$\\
RO$^{800}$+SR$^{800}$ & $0.966$ & $\mathbf{0.995}$ & $0.979$  & $\mathbf{0.986}$  & $\mathbf{0.981}$\\
RO$^{800}$+SR$^{1600}$ & $\mathbf{0.970}$ &  $0.994$ & $\mathbf{0.987}$  & $0.987$  & $0.985$ \\
\midrule
RO$^{200}$+SR$^{800}$ & $0.962$ & $0.992$ & $0.952$  & $0.978$  & $0.971$\\
RO$^{400}$+SR$^{800}$ & $0.965$ & $0.992$ & $\mathbf{0.979}$  & $0.978$  & $0.979$\\
RO$^{800}$+SR$^{800}$ & $0.966$ & $\mathbf{0.995}$ & $\mathbf{0.979}$  & $0.986$  & $0.981$\\
RO$^{1600}$+SR$^{800}$ & $\mathbf{0.968}$ & $\mathbf{0.995}$ & $0.978$ & $\mathbf{0.992}$ & $\mathbf{0.983}$\\
\bottomrule
\end{tabular}
  \begin{tablenotes}
  \item[Note 1] Largest AP and mAP values are highlighted in bold.
  \item[Note 2] We used the following hyper-parameter setting $t_{blue}=0.4$ \& $t_{others}=0.15$, $T=30$ to collect the SB data sets. The values were the same as the experiments in Section V.A.
  \end{tablenotes}
\end{threeparttable}
\end{table}


\subsubsection{Influence of hyperparameters}
Besides the data combination ratio, we also studied the influence of pasting number $T$ and occlusion threshold $t$ used in the SR method. We set both the RO and SR image numbers to $800$ and observed the performance of detectors trained with data sets collected using different $T$ and $t$ values.
Although we previously used a different $t$ value for the ``Blue Tube'', we did not differentiate the tubes here. Like the study on different data combination ratios, this study also comprised two parts. In the first part, we fixed $t$ to be $0.1$ and increased $T$ from $10$ to $40$ with a step length of 10. The upper section of Table \ref{tab:T_t} shows the precision changes under the parameter variations. The results exhibited a significant increase from $10$ to $30$. However, an even larger $T$ had little influence on the recognition performance. In the second part, we set $T$ to be $30$ and varied $t$ from $0.20$ to $0.80$ with a step length of $0.2$. The lower section of Table \ref{tab:T_t} shows the precision changes under the parameter variations. The results exhibited a clear precision increase on the ``Blue Tube''. We speculate that the reason was that the ``Blue Tube'' was shorter and vulnerable to occlusions. A larger $t$ helped provide more occlusion cases in the training data set, leading to a higher detection rate. The results also indicated that the precision of the "White Tube" and "Purple Ring Tube" irregularly changed as the $t$ increased. They were taller and did not suffer from occlusions. Adding occlusions for them caused unexpected errors. For a complete observation, we recommend interested readers to compare with the third row of the table's upper section to catch the changes starting from $t=0.1$. The $T$ value of the upper section's third row was the same as the rows in the lower section.

\begin{table}[htbp]
\caption{Influence of parameters used for synthesis} 
\label{tab:T_t}
\centering 
\begin{threeparttable}
\begin{tabular}{l|cccc|c}
\toprule  
& \multicolumn{4}{c|}{AP} &\\
\cmidrule{2-5}
Params. ($T$, $t$) & Blue &  Purple & White &  Purple Ring  &  mAP \\
\midrule
(10, 0.10) & $0.904$ & $\mathbf{0.995}$ & $0.971$  & $0.973$ & $0.961$\\
(20, 0.10) & $0.915$ & $\mathbf{0.995}$ & $0.976$ & $0.985$ & $0.968$\\
(30, 0.10) & $\mathbf{0.939}$ & $\mathbf{0.995}$ & $\mathbf{0.987}$  & $\mathbf{0.992}$  & $\mathbf{0.978}$\\
(40, 0.10) & $0.934$ & $0.994$ & $ 0.972$ & $0.983 $ & $0.970$\\
\midrule
(30, 0.20) & $0.945$ & $ \mathbf{0.995}$ & $\mathbf{0.985}$  & $0.989$  & $0.978$\\
(30, 0.40) & $0.969$ & $\mathbf{0.995}$ & $\mathbf{0.985}$  & $\mathbf{0.994}$  & $0.986$\\
(30, 0.60) & $0.985$ & $\mathbf{0.995}$ & $0.965$  & $0.967$  & $0.978$\\
(30, 0.80) & $\mathbf{0.987}$ & $\mathbf{0.995}$ & $0.984$  & $0.988$  & $\mathbf{0.988}$\\
\bottomrule
\end{tabular}
\begin{tablenotes}
\item[*] Largest AP and mAP values are highlighted in bold.
\end{tablenotes}
\end{threeparttable}
\end{table}

\subsection{Further Analysis on Synthetic Data} 
We also studied the influence of cap variation and combination ratio on synthetic data sets (the data sets collected using the SR, SB, or SR+SB methods). The goal was to understand the best performance we could reach with synthesis. 

First, we fixed the number of images collected by the SR and SB methods to $800$, respectively. We changed the number of cap region pictures (equals to the number of observation images multiplied by two) used for synthesis from $400$ to $3200$ in a 2-fold ratio to study the influence of cap variation. The previsions YOLOv5x detectors using the changing data sets are shown in Table \ref{tab:abs_R_P}. The results showed that the $400$ row had competitive precision compared to the $1600$ or $3200$ rows. The number was enough to support a satisfying detector. The cap variations were thus considered to have a low influence on learning. 

Second, we fix the number of cap region pictures to $3200$ and change the number of images collected using the SR and SB methods, respectively, to study the influence of the combination ratio. Like the ablation study in Section \ref{subsec:blended_img_num}, we divided the experiment here into two parts. In the first part, we set the number of images collected by the SR method to $800$ and varied the number of images collected by the SB method from $200$ to $1600$ in a 2-fold ratio to understand the importance of the SB data. The upper section of Table \ref{tab:SR_SB_Ratio} shows the precision of detectors trained using the varied data. The number of SB images did not appear to be positively correlated with the final detector's precision, although the largest mAP was observed when the number of SB images was $800$. In the second part, we fixed the number of images collected by the SB method to $800$ and varied the number of images collected by the SR method to understand the importance of the SR data. The lower section of Table \ref{tab:SR_SB_Ratio} shows the precision of detectors trained using the varied data. The result indicated that the mAP improved as the SR image number increased to $800$. There was no significant difference when the image number increased from $800$ to $1600$.

\begin{table}[htbp]
\caption{The influence of \textit{\#}caps to synthesis} 
\label{tab:abs_R_P}
\centering 
\begin{tabular}{l|cccc|c}
\toprule  
& \multicolumn{4}{c|}{AP} &\\
\cmidrule{2-5}
\textit{\#}Caps & Blue &  Purple & White & Purple Ring  &  mAP \\
\midrule
$400$  & $0.970$ & $0.994$ & $0.969$  & $0.984$ & $0.979$\\
$800$  & $0.971$ & $0.993$ & $0.954$ & $0.976$ & $0.973$\\
$1600$  & $0.971$ & $0.992$ & $0.980$ & $0.985$ & $0.982$\\
$3200$ & $0.973 $ & $0.993$ & $0.969$  & $0.985$  & $0.980$\\  
\bottomrule
\end{tabular}  
\end{table}

\begin{table}[htbp]
\caption{Influence of the SR and SB ratio 
} 
\label{tab:SR_SB_Ratio}
\centering 
\begin{threeparttable}
\begin{tabular}{l|cccc|c}
\toprule  
& \multicolumn{4}{c|}{AP} &\\
\cmidrule{2-5}
Data Set Names & Blue &  Purple & White &  Purple Ring  &  mAP \\
\midrule
SB$^{200}$ + SR$^{800}$  & $\mathbf{0.975}$ & $\mathbf{0.994}$ & $0.943$  & $0.957$ & $0.967$\\
SB$^{400}$ + SR$^{800}$ & $0.973$ & $0.990$ & $0.960$ & $0.970$ & $0.973$\\
SB$^{600}$ + SR$^{800}$ & $0.967$ & $0.988$ & $0.951$ & $0.968$ & $0.969$\\
SB$^{800}$ + SR$^{800}$ & $0.973 $ & $0.993$ & $\mathbf{0.969}$  & $\mathbf{0.985}$  & $\mathbf{0.980}$\\
SB$^{1600}$ + SR$^{800}$ & $0.952$ & $0.978$ & $0.926$ & $0.972$ & $0.957$\\
\midrule
SB$^{800}$ + SR$^{200}$ & $0.951$ & $0.982$ & $0.925$ & $0.963$ & $0.955$\\
SB$^{800}$ + SR$^{400}$ & $0.932 $ & $0.969$ & $0.914$ & $0.952$ & $0.942$\\
SB$^{800}$ + SR$^{600}$ & $0.966$ & $0.990$ & $0.930$ & $0.893$ & $0.945$\\
SB$^{800}$ + SR$^{800}$ & $0.973 $ & $\mathbf{0.993}$ & $\mathbf{0.969}$  & $0.985$  & $\mathbf{0.980}$\\
SB$^{800}$ + SR$^{1600}$ & $\mathbf{0.975}$ & $ \mathbf{0.993}$ & $0.967$ & $\mathbf{0.986}$ & $\mathbf{0.980} $\\
\bottomrule
\end{tabular}
  \begin{tablenotes}
  \item[Note 1] Largest AP and mAP values are highlighted in bold.
  \item[Note 2] We used the following hyper-parameter setting $t_{blue}=0.4$ \& $t_{others}=0.15$, $T=30$ to collect the SB data sets, and used the following hyper-parameter setting $T=30$, $t=0.15$ (same for all tubes) to collect the SR data sets. The values were the same as the experiments in Section V.A.
  \item[Note 2] We used 3200 segmented cap region pictures for both methods.
  \end{tablenotes}
\end{threeparttable}
\end{table}

\section{Conclusions}
\label{sec:conclusions}
In this paper, we proposed an integrated robot observation and data synthesis framework for data preparation. The proposed framework can significantly reduce the human effort in data preparation. It required only a single process and was a low-cost way to produce the combined data. The experimental result showed that combined observation and synthetic images led to comparable performance to manual data preparation. The ablation studies provided a good guide on optimizing data configurations and parameter settings for training detectors using the combined data. 


\bibliographystyle{IEEEtran}
\bibliography{citations}

\end{document}